# Towards Bio-inspired Heuristically Accelerated Reinforcement Learning for Adaptive Underwater Multi-Agents Behaviour

[1,3,4]*Antoine Vivien, A.V., [1]Thomas Chaffre, T.C., [1,3]Matthew Stephenson, [4]Eva Artusi, E.A., [1,3]Paulo E Santos, P.E.S., [1,2,3]Benoit Clement, B.C., & [1,3]Karl Sammut, K.S.

*lead presenter

[1] vivi0037@flinders.edu.au, Flinders University, Centre for Defence Engineering, Research and Training, College of Science and Engineering, Adelaide, Australia

[2] ENSTA Bretagne, Brest, France

[3] CROSSING IRL CNRS 2010, Adelaide, Australia

[4] Naval Group, Ollioules, France

This paper describes the problem of coordination and collaboration of an autonomous Multi-Agent System (MAS) which aims to solve the coverage planning problem in a complex environment. The considered applications are the detection and identification of objects of interest while covering an area. These tasks, which are highly relevant for space applications, are also of interest among various domains including the underwater context, which is the focus of this study. In this context, coverage planning is traditionally modelled as a Markov Decision Process (MDP) where a coordinated MAS, i.e. a swarm of heterogeneous autonomous underwater vehicles, is required to survey an area and search for objects. This MDP is associated with several challenges: environment uncertainties, communication constraints, and an ensemble of hazards, including time-varying and unpredictable changes in the underwater environment. Multi-Agent Reinforcement Learning (MARL) algorithms can solve highly non-linear problems using deep neural networks and display great scalability against an increased number of agents. Nevertheless, most of the current results in the underwater domain are limited to simulation due to the high learning time of MARL algorithms. For this reason, a novel strategy is introduced to accelerate this convergence rate by incorporating biologically inspired heuristics to guide the policy during training. The Particle Swarm Optimization (PSO) method, which is inspired by the behaviour of a group of animals, is selected as a heuristic. It allows the policy to explore the highest quality regions of the action and state spaces, from the beginning of the training, optimizing the exploration/exploitation trade-off. The resulting agent requires fewer interactions to reach optimal performance. The method is applied to the Multi-Agent Soft Actor-Critic (MASAC) algorithm and evaluated for a 2D covering area mission in a continuous control environment. The results demonstrate that the proposed method reduces the training time.

# 1. Introduction

Coverage planning is an important task that is required in various environments. It refers to the process of determining a path or set of movements, that ensures a robot or swarm thoroughly covers a given area or surface. This is a fundamental problem in robotics, especially for tasks requiring complete or efficient space exploration, such as surveying, detection, inspection, or search-and-rescue missions. The goal is to systematically minimize unexplored parts of the area, and multi-agents ensure higher efficiency in the operation by combining coordination and collaboration between each other.

In a swarm of agents, coordination means adjusting individual actions so they fit well with what others are doing, and collaboration means teaming up and working together to achieve a common goal. Fully autonomous systems address several human-driven system limitations such as safety operation, robustness, multi-agent coordination, adaptability in complex environments, and energy management. However, these systems are still limited to simulations due to several difficulties in implementing them in the real world.

These difficulties include environmental hazards and complexities such as low-bandwidth communication constraints with short-range, time-varying and unpredictable environmental state changes, and environmental uncertainties.

In particular, the underwater domain has seen a significant uptake in unmanned vehicle development where most of these vehicles are remotely operated rather than fully autonomous. Some encountered challenges are similar to underwater and space environments. These challenges include complex 3D environments with important terrain variations, energy constraints, partial observability, and dynamics including underwater sea currents or space magnetic fields that can endanger the system. For space missions like planetary exploration, multiple rovers, drones, or satellites must work together to map or explore vast regions of a planet, moon, or asteroid. Like underwater agents, these space agents must self-coordinate and cover different sections of the surface or atmosphere. All these challenges [1] make the operations difficult to coordinate across swarms of autonomous agents.

Thus, the focus of this project is on multiple Autonomous Underwater Vehicles (AUVs) working as a team to survey the underwater domain, searching for objects of interest [2]. This involves considering a fully autonomous Multi-Agent System (MAS) that aims to solve the coverage planning task requiring coordination and collaboration in a complex environment with limited communication.

Recent works [3] showed that classical methods (e.g. cell decomposition, potential field, visibility-based) are still lacking efficiency and robustness to handle all challenges because they are not well-suited for complex environments. Thus, researchers have worked and developed important advancements using machine learning methods such as Reinforcement Learning (RL) algorithms [4]. In particular, Multi-Agent Reinforcement Learning (MARL) [5] presents state-of-the-art performances among the different approaches within this context. Indeed, using deep neural networks, MARL allows each agent to learn how to autonomously evolve in complex environments such as underwater and planetary exploration, handle uncertainties and high-dimensional continuous spaces, and enable coordination and collaboration for MAS by allowing decentralized decision-making [6]. However, MARL algorithms introduce a low sample efficiency due to a huge number of samples required to reach a certain level of performance, and a long training time due to a high number of agents and high-dimensional spaces. These constraints limit most of the results to simulation when trying to learn in real-time [7]. In the face of these difficulties, a novel method is studied by incorporating biologically-inspired heuristics [8] to reduce the learning time. The objective is to use heuristics biologically inspired by the behaviour of groups of humans or animals [9,10] to accelerate speed convergence and facilitate an embedded learning system that could learn in real-time. This method improves the exploration by guiding the RL agent towards promising space regions, leading to high-quality rewards.

This paper introduces related works in the second section separated into two parts: MARL and Biologically-inspired Methods. Then, the third section presents the research method including the created customizable environment, the proposed novel method, and the real-world scenario. The evaluation protocol explains the results in section four. Finally, section five discusses future works and improvements and the conclusion is in section six.

## 2. Related Works

### 2.1 Multi-Agent Reinforcement Learning

In Reinforcement Learning (RL) [4], an environment is modelled as a Markov Decision Process (MDP) where an agent learns to make decisions by experiencing different states, taking actions and receiving rewards. This environment can be either known or unknown. Thus, RL allows an agent to learn through trial and error, as it doesn't initially know how the environment behaves. The goal for the agent is to learn a policy (a strategy) that maximizes cumulative rewards over time by telling it which actions to take in each state. The policy represents a mapping from states of the environment to actions. This corresponds to solving the MDP by learning the optimal policy. However, depending on the complexity of the process, an important amount of experience could be required for the policy to perform better than randomness [11]. This is even more critical when considering multi-agent systems due to the increasing dimension of action and state spaces [5].

MARL [12] is a class of solution methods that aims to develop algorithms capable of simultaneously training multiple agents operating in a shared environment. Within this framework, each agent's action and state spaces are uncorrelated. Therefore, the learning agent's objective is to maximise its individual reward while considering the other agents' behaviour and returns (e.g. agent sensor outputs).

Even though MARL methods have good generalization and adaptability as they can perform well in unseen states, one of the key challenges is to consider the different interactions between agents through coordination, competition or cooperation [13]. It includes difficulties in complex environments with high-dimensional continuous spaces and non-stationarity increasing with the number of agents, and dynamic and unpredictable interactions among them [14].

This work assumes the more recent approach of fully decentralised learning [15], where agents learn independently without a central coordinator [16]. These methods pose significant challenges in environment exploration, agents' coordination, and convergence but allow for scalability and robustness in large-scale systems with many agents [17]. Then, several factors need to be considered: the number of agents that will influence the level of communications (limited in bandwidth and range) among them and the covered area, and the time spent on the learning process due to the non-structural, dynamic, and unknown underwater area.

In the context of AUVs, the design of MARL methods is associated with the two following main challenges. First, the multi-agent robustness in the AUV context, which results in maintaining coordination and collaboration of the system, is difficult due to the high uncertainties of the underwater environment and its dynamic nature. This includes unknown dynamics (under the form of disturbing forces acting on the AUV body) and non-linearities (as the operating conditions evolve throughout the mission, the control system is required to be overly conservative). Secondly, current RL algorithms (e.g. Q-learning, Deep Deterministic Policy Gradient) are not suitable for underwater multi-agent contexts due to their low sample efficiency and their tendency to often choose actions in a small area around the current best policy, making it easy to converge to local minima or to overlook insightful regions of the state and action spaces [18].

Moreover, some constraints are related to the application of RL on physical robotic systems [7] including the tracking of high-dimensional continuous state and action spaces and the search for efficient solutions to multi-objective reward functions. It is also important to define the challenges to maintain a suitable level of safety and robustness of the system throughout the mission.

Following these constraints and challenges, Deep Reinforcement Learning (DRL) algorithms [4] are more promising as they combine neural networks with RL to learn optimal policies from high-dimensional spaces and present many advantages. When provided with enough interactions, DRL algorithms can presumably reach near-optimal policy in such contexts. They can learn in complex environments, handle uncertainties with high-dimensional continuous spaces, and enable multi-agent coordination and collaboration [19].

Among the different DRL methods, the focus has been placed on the actor-critic methods [20] which combine value-based and policy-gradient methods, using an actor to propose actions and a critic to evaluate them [21]. Value-based methods [22] estimate the value of each state or state-action pair to derive optimal policies by selecting actions that maximize the cumulative reward. Policy-gradient methods [23] directly optimize the policy by adjusting the parameters of a stochastic or deterministic policy to maximize expected rewards, using gradient descent.

Using a combination of these two approaches, actor-critic methods present some advantages such as good stability with reduced variance, an improved generalisation, and a better trade-off between exploration and exploitation [24] thanks to the stochastic policies [25]. Thus, actor-critic methods are also well-suited for high-dimensional continuous spaces. However, these methods have slow convergence speed due to exhaustive methods and various agents, and low sample efficiency as they require a huge amount of training data to reach a suitable performance level. Within this context, one particular area has not been fully explored yet, the application of a biologically inspired heuristic to accelerate MARL, which is the main interest of this work.

### 2.2 Biologically-inspired Heuristics

Heuristics [26,27] are domain-specific rules or strategies that guide the agent's decision-making process on prior knowledge or expert insights. It can be biologically inspired when the behaviour of groups of humans or animals inspires it. Different classes of bio-inspired methods can be used as a guide for the agent's decision-making process to promising space regions, leading to high-quality rewards. Thus, it falls under the umbrella of Heuristically Accelerated Reinforcement Learning (HARL) [28,29] for MARL, which is a set of methods that combine RL with heuristics to improve learning efficiency, overall performances, and the balance in the exploitation-exploration trade-off. Exploitation means that the agent chooses the action that it believes is the best based on its current knowledge (i.e., the action that gives the highest reward), and exploration means that the agent tries out new actions that it has not taken much or at all, to discover potentially better actions for the future.

Swarm intelligence and social cognitive algorithms are very popular sets of methods for the considered application. Swarm intelligence algorithms are a class of bio-inspired algorithms inspired by the collective behaviour of decentralized, self-organizing systems, such as flocks of birds, schools of fish, or colonies of ants. Some algorithms from this class include the Ant Colony Optimization [30], Bee algorithm [31], and Firefly algorithm [32]. Then Social cognitive algorithms are inspired by human cognition and social behaviour, and by how individuals in a group learn from each other and make decisions based on social interactions. Brain Storm Optimization [33], Group Search Optimization [34], and Social Cognitive Optimization [35] are probably the most famous algorithms of this class. These algorithms possess good performances in solving optimization problems, coordination and collaboration tasks, and decision-making challenges in MAS, and one particular Swarm Intelligence algorithm, namely, Particle Swarm Optimization (PSO) looks interesting because of its simplicity of use and implementation.

PSO [36] is a computational method used to solve optimization problems by iteratively improving a candidate solution concerning a given measure of fitness. The fitness function quantifies the quality of a solution based on the problem's objectives and constraints. The method consists of a group of particles moving around in the search space and each of them represents a potential solution to the optimization problem. Particles adjust their positions based on their own experience and the experience of neighbouring particles, they all know both their best individual position so far and the best position found by the swarm.
Depending on the problem, the goal can be to either maximize or minimize the fitness value. PSO presents many advantages such as a simple implementation because it doesn't require gradient information, good flexibility and adaptability as it can handle non-linear and multi-dimensional optimization problems with multiple constraints and objectives, and it requires tuning of fewer parameters compared to other optimization algorithms [37,38].

## 3. Methodology

### 3.1 A novel strategy to MARL coverage planning

The research method in the domain of MARL started with a benchmark of the different suitable methods for coverage path planning, including a comparison between some recent MARL algorithms and more precisely, some recent actor-critic methods Among the recent actor-critic methods for underwater applications [39], the first version of the SAC [20] in a MAS, namely, Multi-Agent Soft Actor-Critic (MASAC) presents state-of-the-art performances and a guarantee of stability.

Then, the epsilon-greedy method is chosen to improve exploration during training and facilitate the incorporation of the PSO with the first version of the MASAC. The epsilon-greedy method [25] is a simple and popular strategy used in RL for balancing exploitation and exploration when an agent is learning how to make decisions.
The regular method uses a probability of epsilon to control the balance between random exploration and choosing the best-known action. It ensures that the agent does not miss out on potentially better actions by sticking only to what it already knows.

Moreover, the learning of each agent will also be helped by using a replay buffer [40]. It is a memory storage used in reinforcement learning to store past experiences, which are usually made up of the state, action, reward, next state, and done (end of the episode). These experiences are saved in the buffer and then randomly sampled during training. This helps the agent learn from past experiences instead of just from the most recent ones. By sampling experiences randomly, the replay buffer also helps prevent the learning process from being biased toward recent experiences and allows the agent to learn more effectively from a diverse set of interactions with the environment.

Thus, our novel strategy involves the implementation of the bio-inspired heuristic, which is the PSO within the epsilon-greedy method, for the defined context. Instead of using random actions with a specific epsilon probability, the objective is to use the solution proposed by the PSO which is the best candidate for the optimization problem. That way, the exploration is improved and accelerates the convergence speed of the agents. The optimization problem (fitness function) is defined in the same way as the reward function of the MARL in order to guarantee this potential acceleration during learning. This ensures that the actions outputted from the PSO will conduct the agents to high-quality rewards during the learning.

### 3.2 Customizable Environment

As a starting point for the first test of our method, a fully customizable environment that tests the MASAC algorithm on a 2-dimensional coverage area mission is created. Thus, all parameters are known in the actor, critic, replay buffer, algorithm and environment.

The current objective of this environment is coverage planning and detection of objects of interest performed by a team of 3 agents [2]. The team is considered to be homogeneous, meaning that all agents have the same physical properties and functionalities including the state and action spaces. The action space is composed of two values normalized between 0 and 1: one coefficient acting on the agent's linear speed, and one acting on the agent's angular speed. For each agent, the state space includes its position, angle, the last generated actions, other agents' positions, and the updated map. This map is defined to be of size 30 by 30 and each cell is considered to represent one pixel, composing 900 pixels in total that need to be covered by the team of 3 agents. Considering that one pixel is a unit, the agent's maximum linear speed is set as 5 units per step and the agent's maximum angular speed is 30 degrees per step.

Moreover, in order to respect reality, each agent simulates a looking-forward sensor by having their covering view in the same form at each step. This is why this covering view has been defined as a section of a circle with an angle of 60 degrees and a radius of 2 units. Regarding this definition, at each step, the maximum number of new possible pixels covered by an agent is 7.

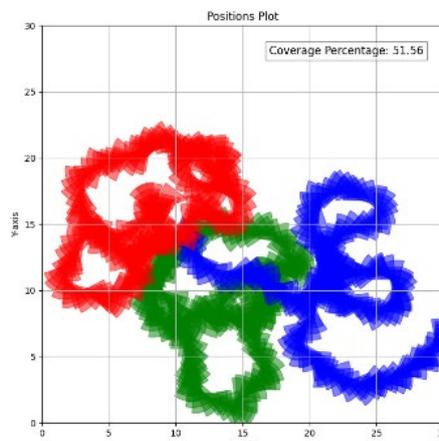

*Figure 1 Customizable environment for coverage planning with 3 agents.*

The reward function for each agent is defined as follows:

$$r_t \begin{cases} -5 \text{ if collision with an agent} \\ -5 \text{ if out of bounds} \\ +1 \text{ for each new covered pixel} \end{cases}$$

Regarding the PSO parameters, the first results shown in section four have been obtained with 50 particles (possible solutions), 100 iterations, and 50 time-steps. The iterations determine how many times the swarm of particles will update their positions to converge towards the best solution while the time steps represent the discrete time intervals over which the agents move and interact with the environment (map) to cover areas. Other parameters have been defined following a modifier PSO in [37]. Thus, the inertia value is set at 0.8, and both cognitive and social components are at 2.0.

### 3.3 Scenario

The proposed scenario is a Mine Counter Measure (MCM) mission. It consists of a group of AUVs that aim to detect Mine-Like Objects (MLOs) buried under, sitting on, or suspended above the seabed [2]. For this purpose, a scenario is defined. It exploits hybrid teams of AUVs including scanning sonar-equipped torpedo-shaped vehicles to scout ahead scanning for possible MLOs followed by hovering type vehicles equipped with different sensing modalities, such as cameras, that are then tasked to inspect and confirm potential MLOs close up. The scenario is designed to ensure that it can be simulated and transferred in the real world using the robotic platforms available at Flinders University. It is also designed to incorporate operational constraints that will be encountered in the real world such as sensors and communication limitations, battery life, and system energy consumption.

### 4. Results

When the learning begins, a starting point is defined at each episode from which all the agents start one beside the others. One of the chosen metrics for the evaluation is the reward value. The desired result is first, an increasing value along the training, and second, higher values with the PSO compared to values obtained without it. The shown results present promising values as they are all higher and reach better performances faster than without the PSO. This proves that the learning performs better in fewer episodes as it can reach better mean rewards, which is associated with better covering of the map.

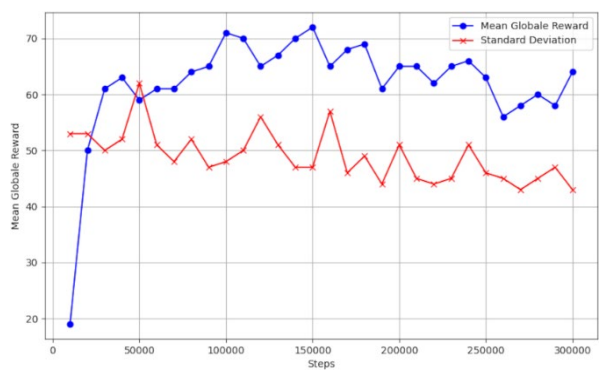

*Figure 2 Training performance without PSO.*

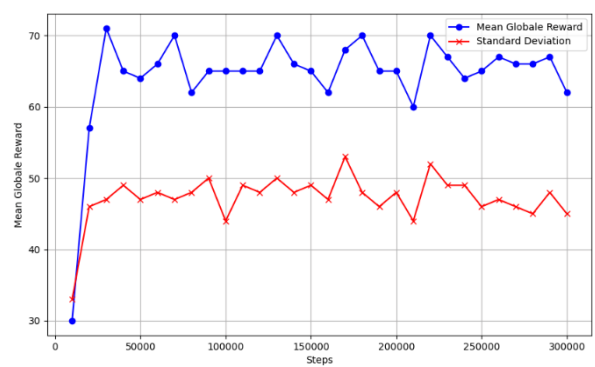

*Figure 3 Training performance with PSO.*

These preliminary results support the hypothesis that the addition of heuristics within the exploration strategy leads to lower training time. This can be explained by the fact that the actions resulting from the PSO are on average associated with higher reward values, which would otherwise be experienced only later during the training without heuristics. With the heuristic, the MASAC algorithm can exploit such actions from the beginning of the training to perform the Policy Gradient update. The benefits of this approach are explained by the fact that Policy Gradient methods do not learn what good actions are, but rather compare multiple actions to each other and adjust their associated probability to be taken accordingly (through the Advantage function [4]). In this context, providing the agent with actions associated with a reward value higher than average allows instant improvement of the agent's return.

## 5. Conclusion

This paper has described the problems related to the coordination and collaboration of an autonomous MAS performing coverage planning in a complex environment, particularly in the underwater context of detecting and identifying objects of interest while surveying an area. Key challenges include managing interactions between agents, ensuring reliability with communication constraints and environmental disturbances, and improving the efficiency of MARL algorithms.

To address the problem of high learning time, a novel strategy has been introduced with the implementation of a bio-inspired heuristic, the PSO, in an epsilon-greedy method, to guide the policy towards promising regions of the action-state spaces to optimize the exploration/exploitation trade-off. It demonstrated that the agent experiences higher-quality actions early in training, requiring fewer interactions to reach better performances.

Despite the promising performances of the first version of the MASAC, it will be pertinent to switch to its latest version, namely, the SAC with Automatic Entropy Adjustment [41]. This algorithm possesses better performances compared to the first version, however, due to the lack of convergence guarantee on the dual policy update [41] it can be challenging to stabilize its training. Thus, an interesting work would be to find a novel approach to improve this stability. The latest MASAC also requires additional hyperparameters tuning to reach optimal performances. The PSO can also be optimized by adjusting some hyperparameters such as its number of iterations, particles, inertia, and cognitive and social components. Some works have already demonstrated that DRL can be used to determine the optimal hyperparameters of the PSO algorithm. Therefore, automatic tuning of the PSO could be considered in the future.

Moreover, other future works will involve the reduction of the state vector dimension. Little to no loss of information could be achieve in the process by using deep learning methods (e.g. Convolutional Neural Networks, Recurrent Neural Networks, Generative Adversarial Networks) or model-based methods (e.g. Singular Value Decomposition, Principal Component Analysis, Independent Component Analysis). A reduced state vector will likely improve the learning process by reducing variance, leading to lower training time and better performance.

In addition to the direct exploration method proposed in this paper, passive exploration, as controlled by the Experience Replay (ER) mechanism, could also reduce the training time. Depending on the size of the Replay Buffer, a high-quality interaction, performed by following the PSO algorithm, might take a long time before being sampled in the ER process and affecting the current policy [40]. Recently, there have been promising results in optimizing which transition to use in the experience replay mechanism based on the reward value [42].